\documentclass[10pt,twocolumn,letterpaper]{article}
\usepackage[accsupp]{axessibility}  
\usepackage{cvpr}              
\usepackage{cuted}
\usepackage{capt-of}
\usepackage{graphicx}
\usepackage{amsmath}
\usepackage{amssymb}
\usepackage{booktabs}
\usepackage{floatrow}
\usepackage{multicol,graphicx,mathrsfs,pifont,amscd,latexsym,color,url, pifont, epstopdf,setspace,multirow,colortbl, bbm,listings, balance,color,indentfirst,enumitem, amsmath,amssymb, verbatim,microtype,xcolor}
\usepackage{graphicx,booktabs, paralist}

\usepackage{algorithm}
\usepackage{algorithmic}

\newfloatcommand{capbtabbox}{table}[][\FBwidth]

\usepackage[pagebackref,breaklinks,colorlinks]{hyperref}

\usepackage{algorithm}
\usepackage{algorithmic}

\usepackage{fancyhdr}

\newcommand{\ignore}[1]{}

\usepackage[capitalize]{cleveref}
\crefname{section}{Sec.}{Secs.}
\Crefname{section}{Section}{Sections}
\Crefname{table}{Table}{Tables}
\crefname{table}{Tab.}{Tabs.}

\def\cvprPaperID{3930} 
\def\confName{CVPR}
\def\confYear{2023}
\newcommand{\method}{R\textsc{e}V\textsc{ea}L\xspace}

\begin{document}


\author{%
   Ziniu Hu$^1$\thanks{This work was done when Ziniu was an intern at Google.}\
   ,\ \ Ahmet Iscen$^2$,\ \ Chen Sun$^2$,\ \ Zirui Wang$^2$,\ \ Kai-Wei Chang$^1$,\ \ Yizhou Sun$^1$ \\ 
   Cordelia Schmid$^2$,\ \ David A. Ross$^2$,\ \ Alireza Fathi$^2$  \\
      $^1$University of California, Los Angeles,\ $^2$Google Research
}


\title{\method: Retrieval-Augmented Visual-Language  Pre-Training with \\Multi-Source Multimodal Knowledge Memory}
\maketitle
\begin{strip}\centering
~\vspace{-4em}\\
    \includegraphics[width=1.8\columnwidth]{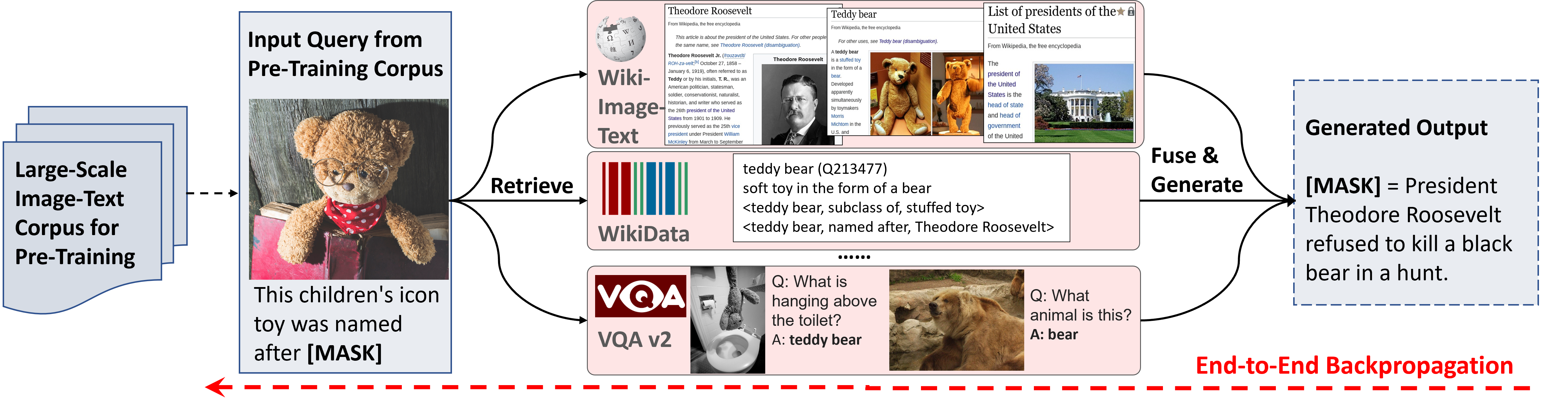}
~\vspace{-1em}\\    
    \captionof{figure}{We augment a visual-language model with the ability to retrieve multiple knowledge entries from a diverse set of knowledge sources, which helps generation. Both retriever and generator are trained jointly, end-to-end, by optimizing a language modeling objective.}
    \label{fig:fusion}
\end{strip}

\begin{abstract}


In this paper, we propose an end-to-end \underline{Re}trieval-Augmented \underline{Vi}sual \underline{L}anguage Model (\method) that learns to 
encode world knowledge into a large-scale memory, and to retrieve from it to answer knowledge-intensive queries.
\method consists of four key components: the memory, the encoder, the retriever and the generator. 
The large-scale memory encodes various sources of multimodal world knowledge (\eg image-text pairs, question answering pairs, knowledge graph triplets, \etc) via a unified encoder. 
The retriever finds the most relevant knowledge entries in the memory, and the generator fuses the retrieved knowledge with the input query to produce the output. A key novelty in our approach is that the memory, encoder, retriever and generator are all pre-trained end-to-end on a massive amount of data. Furthermore, our approach can use a diverse set of multimodal knowledge sources, which is shown to result in significant gains.  
We show that \method achieves state-of-the-art results on visual question answering and image captioning. The project page is {\tt \href{ReVeaL-CVPR.github.io}{\textbf{\textcolor{orange}{ReVeaL-CVPR.github.io}}}}.

\end{abstract}

\everypar{\looseness=-1}
\section{Introduction}\label{sec:introduction}


Recent large-scale models such as T5~\cite{DBLP:journals/jmlr/RaffelSRLNMZLL20}, GPT-3~\cite{DBLP:conf/nips/BrownMRSKDNSSAA20}, PaLM~\cite{DBLP:journals/corr/abs-2204-02311}, CoCa~\cite{DBLP:journals/corr/abs-2205-01917}, Flamingo~\cite{DBLP:journals/corr/abs-2204-14198},  BEIT-3~\cite{DBLP:journals/corr/abs-2208-10442}
and PaLI~\cite{DBLP:journals/corr/abs-2209-06794} 
have demonstrated the ability to store substantial amounts of world knowledge, when scaled to tens of billions of parameters and trained on vast text and image corpora. These models achieve state-of-the-art results in downstream tasks such as image captioning, visual question answering and open vocabulary recognition. 
%
Yet, these models have a number of drawbacks: (i) they require massive scale, of parameters, data and computation,
and (ii) they need to be re-trained every time the world knowledge is updated.

To address these issues, we adopt a different approach. Instead of statically compiling world knowledge into model weights, we transform the knowledge into a key-value memory through neural representation learning. Our model learns to utilize the memory for answering knowledge-intensive queries. By decoupling the knowledge memorization from reasoning, we enable our model to leverage various external sources of knowledge (\eg, Wikipedia passages and images~\cite{DBLP:conf/sigir/Srinivasan0CBN21}, the WikiData knowledge graph~\cite{DBLP:journals/cacm/VrandecicK14}, Web image-text pairs~\cite{changpinyo2021cc12m} and visual question answering data~\cite{DBLP:journals/ijcv/GoyalKASBP19}). This enables the model parameters to focus on understanding the query and conducting reasoning, rather than being dedicated to memorization. 

Retrieval-augmented models have attracted a fair amount of attention in the fields of NLP~\cite{DBLP:journals/corr/abs-2002-08909, DBLP:journals/corr/abs-2208-03299} and computer vision~\cite{DBLP:conf/cvpr/Long0ANPGBSH22, DBLP:conf/naacl/GuiWH0BG22}. Typically, these models often use a pre-existing single-modality backbone to encode and retrieve information from the knowledge corpus. Such approaches do not leverage all available modalities in the query and knowledge corpora, and hence they might not find the information that is most helpful for generating the model output. A key novelty in our approach is that we encode and store various sources of multimodal world knowledge into a unified memory, which the retriever can access via multimodal query encodings, to find the most relevant information from across complementary sources. Our multimodal memory and retriever are pre-trained end-to-end together together with the rest of the model, on a massive amount of data and using diverse knowledge sources.

A key challenge of pre-training the multimodal retriever end-to-end is the lack of direct supervision. 
There is no ground-truth indicating which knowledge entries are most helpful for answering knowledge-intensive queries. Some of the existing works in NLP~\cite{DBLP:journals/corr/abs-2002-08909,DBLP:conf/nips/LewisPPPKGKLYR020,DBLP:conf/nips/SachanRHDY21F} propose to acquire training signal by assessing the usefulness of each retrieved knowledge entry independently for helping language modelling. 
This approach is inefficient, as it involves estimating hundreds of retrieved knowledge entries independently, and also inaccurate as it discards the dependency between different knowledge entries in the retrieval set.
In contrast, we propose to get this training signal while simultaneously considering multiple retrieved knowledge entries, by introducing an attentive fusion layer that injects retrieval score into the attention calculation procedure. This enables the retrieval module to be differentiable and jointly pre-trained with the rest of the model.

In summary, our key contributions are as follows: 
\begin{compactitem}
\item 
We are the first to propose an end-to-end pre-training paradigm that learns to index into a large-scale memory to solve knowledge-intensive visual-language tasks.
\item
Our method can construct a large-scale memory by encoding various sources of multimodal world knowledge, including Wikipedia passage, web images with alt-text captions, and knowledge graph triplets.
\item
\method achieves state-of-the-art performance on several knowledge-intensive visual question answering and image captioning datasets. Notably on the OKVQA benchmark, \method achieves a new state-of-the-art, 59.1$\%$ accuracy, while using order of magnitude fewer parameters than previous works.
\end{compactitem}

\section{Related Work and Background}\label{sec:related}

\noindent \textbf{Knowledge-based Visual Question Answering.}
To evaluate a model's ability to comprehend multimodal world knowledge not easily inferred from input data, several knowledge-based Visual Question Answering (VQA) datasets have been introduced. KB-VQA~\cite{DBLP:conf/ijcai/WangWSDH17} and FVQA~\cite{DBLP:journals/pami/WangWSDH18} design questions that can be answered by retrieving relevant triplets from domain-specific structured knowledge graphs. OK-VQA~\cite{DBLP:conf/cvpr/MarinoRFM19} improves these datasets by necessitating the use of external knowledge, which goes beyond what can be directly observed in the input images.
More recently, A-OKVQA~\cite{DBLP:journals/corr/abs-2206-01718} offers further improvements to OK-VQA by exclusively selecting questions that demand both external knowledge and commonsense reasoning about the image scenes.
To tackle knowledge-based VQA tasks, many approaches have been proposed to incorporate external knowledge into visual-language models. One line of research uses explicit knowledge from structured knowledge graphs~\cite{DBLP:conf/nips/NarasimhanLS18, DBLP:conf/emnlp/GarderesZAL20, DBLP:journals/corr/abs-2205-11501, DBLP:conf/emnlp/HuX0WYZCS22} or unstructured text corpora~\cite{DBLP:conf/cvpr/MarinoCP0R21, DBLP:conf/emnlp/LuoZBB21, DBLP:conf/aaai/WuLSM22}. The key component for these works is the knowledge retriever. Some works~\cite{DBLP:conf/nips/NarasimhanLS18, DBLP:conf/emnlp/GarderesZAL20, DBLP:conf/cvpr/MarinoCP0R21, DBLP:conf/aaai/WuLSM22} utilize off-the-shelf vision detection models to generate image tags for knowledge retrieval, while others train the retrieval model via distant supervision~\cite{DBLP:conf/emnlp/LuoZBB21} or auxiliary tasks (\eg entity linking)~\cite{DBLP:conf/naacl/GuiWH0BG22}. Another research direction aims to incorporate implicit knowledge from pre-trained Large Language Models, such as GPT-3~\cite{DBLP:conf/nips/BrownMRSKDNSSAA20} or PaLM~\cite{DBLP:journals/corr/abs-2204-02311}. These approaches utilize off-the-shelf image caption models to convert images into text, feed them into a language model, and use the generated text output as augmented knowledge~\cite{DBLP:journals/corr/abs-2109-05014, DBLP:conf/naacl/GuiWH0BG22, DBLP:journals/corr/abs-2206-01201}. 
Our work follows the first direction, augmenting a vision-language model with an explicit knowledge retriever. The main distinction is that we propose an end-to-end training framework to jointly learn the answer generator and retriever, rather than using a fixed or predefined knowledge retrieval.

\paragraph{End-to-End Training of Retrieval-Augmented Models.}
Given the advantage of knowledge retrieval, a key question is how to get learning signal to train the retrieval model. For tasks with annotated retrieval ground-truth, retrieval training can be conducted via standard contrastive learning~\cite{DBLP:journals/corr/abs-2004-04906}. However, most tasks do not provide clear indications of which knowledge entries are relevant for generating answers. To this end, a series of studies have investigated retrieval training using supervision derived from downstream tasks.
REALM~\cite{DBLP:journals/corr/abs-2002-08909} trains a single-document retriever by concatenating each retrieved result with the query, to calculate the final loss independently. A similar approach has been used by EMDR$^2$~\cite{DBLP:conf/nips/SachanRHDY21F} for multi-document retrieval training. FID-KD~\cite{izacard2020distilling} proposes to use the aggregated attention score calculated by the generator as a distillation signal to train the retriever. Atlas~\cite{DBLP:journals/corr/abs-2208-03299} further introduces a perplexity distillation loss and a leave-one-out variant. Our \method proposes to inject the retrieval scores directly into an attentive fusion module, enabling to train the retriever to directly optimize downstream tasks as well as pre-training objectives. 



\section{Method}\label{sec:approach}
\begin{figure*}[ht!]
    \vspace{-2em}
    \centering
    \includegraphics[width=1.0\columnwidth]{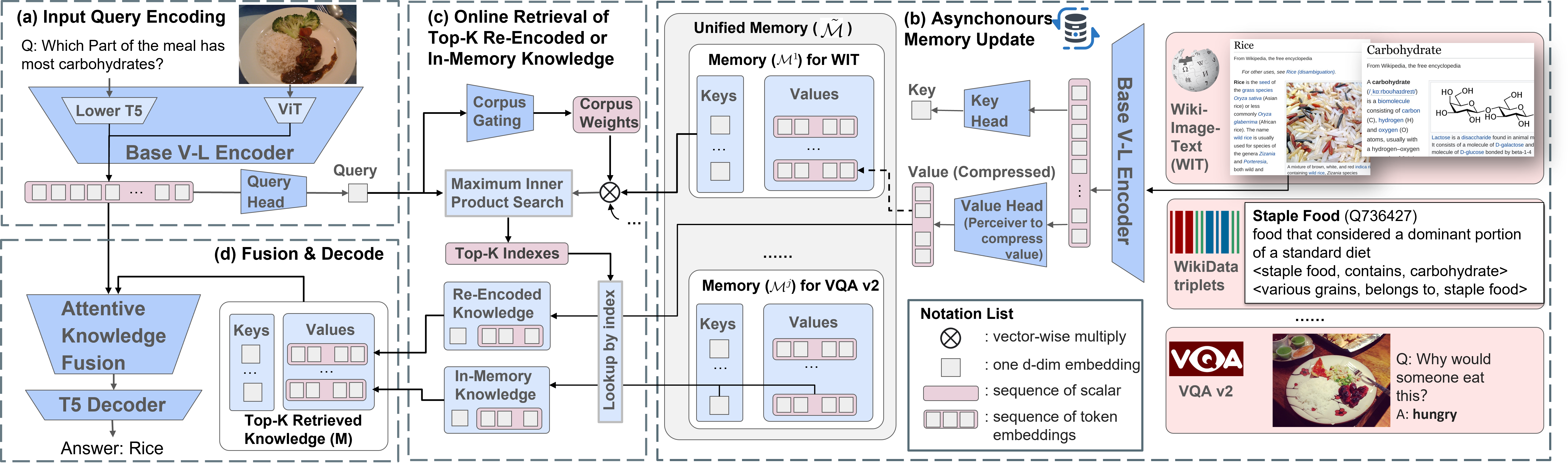}
    \caption{
    \textbf{The overall workflow of \method} consists of four main steps: \textbf{(a)} encode a multimodal input into a sequence of token embeddings and a summarized query embedding; \textbf{(b)} encode each knowledge entry from different corpus into unified key and value embedding pairs, where key is used to index the memory and value contains full information of the knowledge; \textbf{(c)} retrieve top-K 
    most similar knowledge items from different knowledge sources, and return the pre-computed in-memory value embeddings and re-encoded value; 
    and \textbf{(d)} fuse the top-K knowledge items via attentive knowledge fusion layer by injecting the retrieval score as a prior during attention calculation.
    This facilitates \method's key novelty: the memory, encoder, retriever and the generator can be jointly trained in an end-to-end manner. 
    }
    \label{fig:framework}
    \vspace{-.1in}
\end{figure*}



We propose a Retrieval-Augmented Visual Language Model (\method), which learns to use knowledge from different sources for solving knowledge-intensive tasks.
For both pre-training and fine-tuning, our goal is to learn the distribution $P(y \mid x)$ to generate a textual output $y$ conditioned on a multimodal input query $x$. \method contains four components: knowledge encoding, memory, retrieval and generation. 
Given an input query $x$, we first retrieve $K$ possibly helpful entries $M = \{m_1, \cdots, m_K\}$ from the memory corpora $\mathcal{M}$. Each $m$ is a memory entry containing the encoded single key embedding and a sequence of value embeddings (we will describe how to encode knowledge items into memory entries in Sec.~\ref{sec:memory}). With it, the retriever can use embedding similarity to find relevant memory entries.
We model this retrieval process as sampling from distribution $p(M \mid x)$. Then, we condition on both the retrieved set $M$ and the original input query $x$ to generate the output $y$, modeled as $p(y \mid x,M)$. To obtain the overall likelihood of generating $y$, we treat $M$ as a latent variable from the entire memory $\tilde{\mathcal{M}}$ and marginalize over it yielding:
\begin{align}
    p(y  \mid  x) = \sum_{M \subset \tilde{\mathcal{M}}}  \underbrace{p(M  \mid x)}_{retrieval} \cdot  \underbrace{p(y \mid  x, M)}_{generation}. \label{eq:main}
\end{align} 
However, this marginal probability involves an intractable summation over all size-$K$ subsets of the memory corpora $\tilde{\mathcal{M}}$. 
We approximate this instead by using the top-K entries in memory with the highest probability under $p(M \mid x)$. This is reasonable if most of the unrelated memory entries do not contribute to the generation. Note that we use an online memory that is updated as the knowledge encoder is trained end-to-end with the rest of the model.  


Figure~\ref{fig:framework} illustrates the overall workflow of \method, and we describe each component in this section. In particular, in Sec.~\ref{sec:encoding} we describe how the query is encoded. In Sec.~\ref{sec:memory} we go over how the multimodal knowledge memory is constructed and updated during pre-training. Next, we describe how we retrieve the memory entries that are most relevant to the input query in Sec.~\ref{sec:retrieval}. Finally, in Sec.~\ref{sec:generation} we describe the generator that fuses the query and retrieved knowledge and decodes them into the generated text.

\subsection{Query Encoding} \label{sec:encoding}
Figure~\ref{fig:framework} (a) depicts how the input image-text query is encoded.
We use a base visual-language encoder $b(\cdot)$ to turn the query input and each knowledge item (with potentially different modalities \eg text-only, image-only or image-text pairs) into a sequence of embeddings (tokens).
We adopt a Vision Transformer (ViT)~\cite{ViT} to encode the images and we use a lower-layer\footnote{We denote the last $l$ layers of a T5 encoder as `upper-layer', and the remaining ones including the token embedding layer as `lower-layer'.} T5 encoder~\cite{DBLP:journals/jmlr/RaffelSRLNMZLL20} to encode the texts. We add a projection layer on top of the ViT model to map the image tokens into the same space as the text tokens. We then concatenate the two modalities together.
We use an upper-layer T5 module as both the query Head $\phi_{\texttt{Query}}(\cdot)$ and the key Head $\phi_{\texttt{Key}}(\cdot)$ to compute the query embedding and memory keys. We take the output of the first \texttt{[CLS]} tokens followed by a linear projection and 
L2-normalization to summarize the input into a $d$-dimensional embedding.


\subsection{Memory} \label{sec:memory}
Figure~\ref{fig:framework} (b) shows how memory is constructed and updated by encoding knowledge items.
Our approach differs from previous works primarily by leveraging a diverse set of multimodal knowledge corpora (WikiData knowledge graph, Wikimedia passages and images, Web image-text pairs). Throughout the paper, we denote each corpus as $\mathcal{C}^j = \{z^j_1, \dots, z^j_N\}$, in which each $z^j_i \in \mathcal{C}^j$ is a knowledge item that could be an image-text pair, text only, image only, or a knowledge graph triplet. We denote the unified knowledge corpus as $\tilde{\mathcal{C}} = \mathcal{C}^1 \cup \mathcal{C}^2 \dots \cup \mathcal{C}^S$ that combines $|\tilde{\mathcal{C}}|=S$ different knowledge corpora. 
We encode the external knowledge corpora into a unified memory $\tilde{\mathcal{M}}=[\mathcal{M}^1, \dots, \mathcal{M}^{|\tilde{\mathcal{C}|}}]$. Each knowledge item $z_i$ is encoded into a key/value pair $m_i=(\texttt{Emb}_{\texttt{Key}}(z_i),\texttt{Emb}_{\texttt{Value}}(z_i))$ in memory.  Each key
$\texttt{Emb}_{\texttt{Key}}(z)= \phi_{\texttt{Key}}\big(b(z)\big) \in \mathbb{R}^{d}$
is a $d$-dimensional embedding vector encoded via Key Head. 
Each value is a sequence of token embeddings representing the full information of knowledge item $z$.
We follow a similar procedure as in~\cite{DBLP:journals/corr/abs-2002-08909} to precompute key/value embeddings of knowledge items from different sources and index them in a unified knowledge memory.
We continuously re-compute the memory key/value embeddings as the model parameters get updated during the pre-training phase. We update the memory $\tilde{\mathcal{M}}$ asynchronously at every 1000 training steps.

\textbf{Scaling Memory by Compression}
A naive solution for encoding the memory value is to keep the whole sequence of tokens for each knowledge item. Then, the generator could fuse the input query and the top-K retrieved memory values by concatenating all their tokens together and feeding them into a Transformer Encoder-Decoder pipeline~\cite{DBLP:conf/nips/LewisPPPKGKLYR020}. This approach has two issues: (1) storing hundreds of millions of knowledge items in memory is impractical given that each memory value would consist of hundreds of tokens; (2) transformer encoder has quadratic complexity with respect to the total number of tokens times $K$ for self-attention. 

Therefore, we propose to use the Perceiver architecture~\cite{DBLP:conf/icml/JaegleGBVZC21} as the Value Head to encode and compress knowledge items. The Perceiver model uses a transformer decoder $\psi(\cdot)$ with learnable $c$-length latent embeddings to compress the full token sequence into an arbitrary length $c$, such that $\texttt{Emb}_{\texttt{Value}}(z) =\psi(b(z)) \in \mathbb{R}^{c \times d}$ (In our experiments we use $c=32$).
This lets us retrieve top-K memory entries for K as large as a hundred.
To make the compressed embeddings generated by Perceiver more expressive, we add two additional regularizations. The first one is a disentangled regularization~\cite{DBLP:journals/corr/abs-2205-09797} that forces every two output tokens to be linearly de-correlated $\mathcal{L}_{\text{decor}}=\sum_{i,j=1}^K \Big\lVert \text{Covariance} \big(\psi\big(b(z_i)\big), \psi\big(b(z_j)\big)\big)   \Big\rVert_F^2$, and
the second one is an alignment regularization that minimizes the distance of L2-Norm between the query and compressed knowledge embedding: $\mathcal{L}_{\text{align}} = \bigg \lvert 1 -  \frac{\sum_{z} \lVert \psi\big(b(z) \big) \rVert_2}{\sum_{x} \lVert b(x) \rVert_2}  \bigg \rvert$.

\subsection{Retriever} \label{sec:retrieval}
Figure~\ref{fig:framework}~(c) shows \method's retrieval procedure.
Given the input query $x$, the retriever's task is to find top-K memory entries $M$ with the highest probability $p(M \mid x)$ which we approximate as $p(M \mid x) = \prod_{m \in M} p(m \mid x)$ by retrieving each entry independently.
Note that we retrieve from a large-scale unified memory $\tilde{\mathcal{M}}=[\mathcal{M}^1, \dots, \mathcal{M}^{|\tilde{\mathcal{C}|}}]$ that is constructed from a diverse set of knowledge sources. To help the query to better choose the most appropriate knowledge sources, we learn a gating function that models the probability of retrieving from each memory corpus. With the corpus gating, for $m^j_i \in \mathcal{M}^j$ we re-weight $p(m^j \mid x)$ by the computed corpus gating score:
\begin{align}
    &p(m^j_i \mid x)= p(\mathcal{M}^j \mid x) \cdot p(m^j_i \mid x;\mathcal{M}^j)  \label{eq:prob} \\
    &= Gate_{\mathcal{M}^j}(x) \cdot \frac{\exp\big( Rel(x, m^j_i) / \tau \big) }{\sum_{m^{j}_{k} \in \mathcal{M}^j} \exp\big( Rel(x, m^{j}_{k}) / \tau \big)}   \label{eq:retrieve}
\end{align}
where $Gate_{\mathcal{M}^j}(x)=\text{Softmax}(W\cdot \texttt{Emb}_{\texttt{Query}}(x) + b)[j]$ is a softmax gating that assigns a score to each memory corpus $M^j$,
with $W$ and $b$ as function parameters. 
$Rel(x,m^j_i)$ models relevance score between query $x$ and each memory entry via embedding dot product, such that $Rel(x,m^j_i) = \texttt{Emb}_{\texttt{Query}}(x)^T \cdot \texttt{Emb}_{\texttt{Key}}(z^j_i)$. 
where $z_i$ is the knowledge item corresponding to the memory entry $m_i$ and $\tau$ is the temperature parameter.

After identifying the top-K memory entries, the retriever passes the pre-computed in-memory key and value embeddings to the generator. In the meantime, to support end-to-end training of the encoders, we also re-encode a small portion (i.e., 10$\%$) of the retrieved knowledge items $z_i$ from scratch. In this way, the memory encoders could be updated with moderate computational cost.
We concatenate the re-encoded knowledge with in-memory ones to construct the final top-K retrieved key/value embeddings.



\begin{figure}[t!]
    \centering
    \includegraphics[width=1.0\columnwidth]{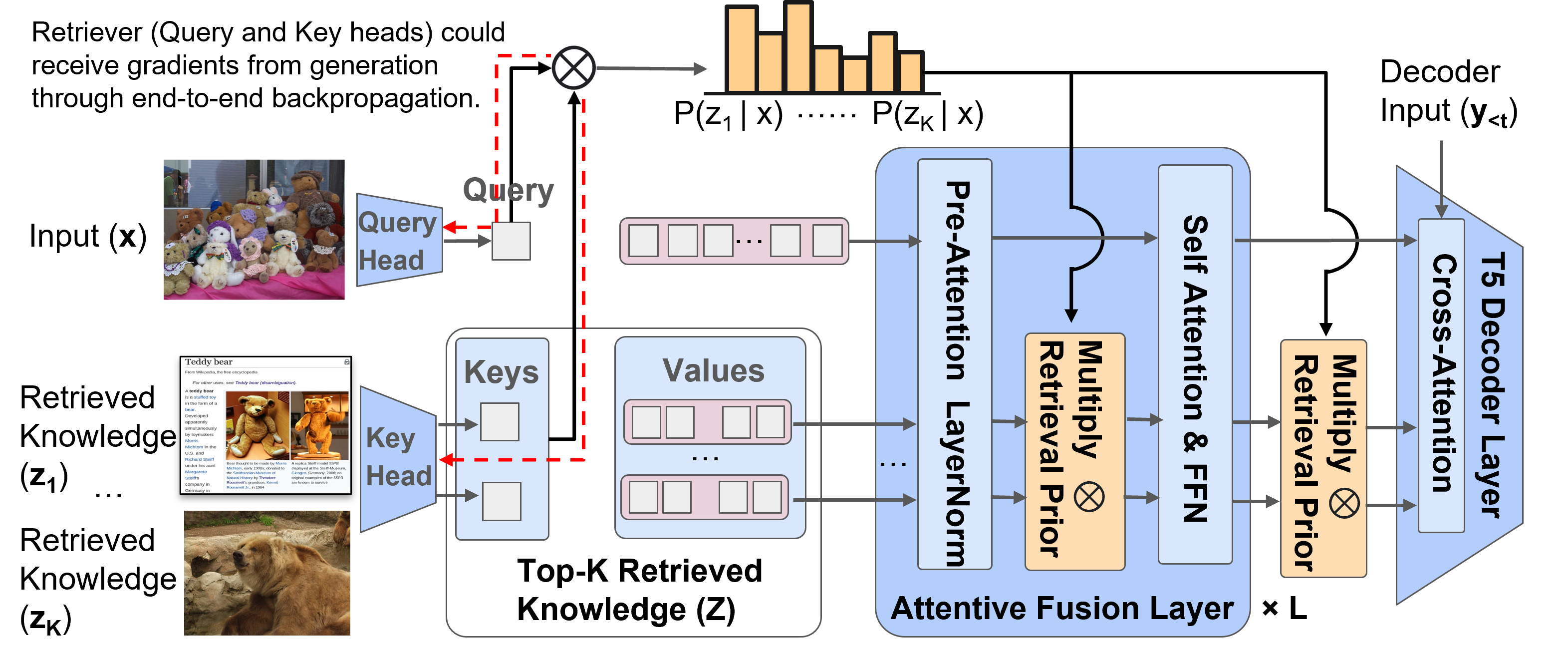}
        \vspace{-.2in}
    \caption{Detailed procedure of attentive knowledge fusion module. We inject retrieval probability as a prior to knowledge token embeddings, so the retriever can receive gradients via back-propagating over \{self/cross\}-attention part.}
    \label{fig:fusion}
\end{figure}

\subsection{Generator} \label{sec:generation}

Figure~\ref{fig:framework}~(d) shows how the query and the retrieved knowledge items are fused to generate the output answer.
All $K$ retrieved memory values are concatenated with the query embedding, which is feasible due to the Perceiver module utilized as the value head $\psi(\cdot)$, compressing each knowledge item into a short sequence. We denote the concatenated query embedding and memory values as $X = [b(x), \psi(b(z_1)), \dots, \psi(b(z_K))] \in \mathbb{R}^{(I + c \cdot K) \times d}$, where $I$ is the number of tokens of the input query $x$ and $c$ is the number of compressed tokens. To guide the generator towards attending to the most important items in $X$ and facilitate backpropagation of gradients to the retriever, we propose an attentive fusion module $f(\cdot)$ capable of incorporating the retriever score as a prior for calculating cross-knowledge attention. The detailed procedure is illustrated in Figure~\ref{fig:fusion}. We firstly 
compute a latent soft attention mask over $X$ as $\texttt{Mask}_{\texttt{att}} = [1,p(z_1|x), \dots, p(z_K|x)]$.
Finally, we pass the fused representation $f(X, \texttt{Mask}_{\texttt{att}})$ into a T5 decoder module $g(\cdot)$ to generate the textual output.
\looseness=-1

\section{Generative Pre-Training}\label{sec:training}
{ 
\setlength\tabcolsep{2pt} 
\begin{table}[t!]
\centering
\footnotesize
\begin{tabular}{c|cccc} \toprule
Knowledge Source & Corpus Size & Type of Text & Avg. Text Length  \\ \midrule
WIT~\cite{DBLP:conf/sigir/Srinivasan0CBN21} & 5,233,186 & Wikipedia Passage & 258 \\
CC12M~\cite{changpinyo2021cc12m} & 10,009,901& Alt-Text Caption & 37  \\ 
VQA-V2~\cite{DBLP:journals/ijcv/GoyalKASBP19} & 123,287 & Question Answer & 111 \\ 
WikiData~\cite{DBLP:journals/cacm/VrandecicK14} & 4,947,397& Linearlized Triplets & 326  \\
\bottomrule
\end{tabular}
\caption{Statistics of the knowledge sources used.}
\label{tab:kb}
\end{table}
}

{ 
\setlength\tabcolsep{2pt} 
\begin{table}[t!]
\centering
\footnotesize
\begin{tabular}{c|ccc|cc} \toprule
Model Name & T5 Variant & Image Encoder &\# params. & GFLOPs \\ \midrule
\method-Base & T5-Base  & ViT-B/16 &  0.4B & 120 \\ 
\method-Large & T5-Large & ViT-L/16  & 1.4B & 528 \\ 
\method & T5-Large & ViT-g/14  &  2.1B & 795\\ \bottomrule
\end{tabular}
    \vspace{-.1in}
\caption{Model configuration of different \method variants.}
\label{tab:variant}
\end{table}
}


The existing VQA datasets are not large enough for training a complex multi-component model like ours from scratch. Therefore, we pre-train our model on a massive image-text corpus. 
In Sec.~\ref{sec:objective} we go over the details of our pre-training data and objective. Then in Sec.~\ref{sec:knowledge-sources} we introduce the various sources of knowledge used in our experiments. Finally, in Sec.~\ref{sec:impl} we describe the pre-training implementation details.
\subsection{Pre-Training Objective} \label{sec:objective}
We pre-train our model on the Web-Image-Text dataset~\cite{DBLP:conf/cvpr/ZhaiWMSK0B22}, a large-scale corpus containing 3 billion image alt-text caption pairs collected from the public Web. Since the dataset is noisy, we add a filter to remove data points whose captions are shorter than 50 characters. This yields roughly 1.3 billion image caption pairs for pre-training.

We denote the pre-training Web-Image-Text dataset~\cite{DBLP:conf/cvpr/ZhaiWMSK0B22} as $\mathcal{D}$. We use the text generation objective used in Wang et al.~\cite{DBLP:conf/iclr/WangYYDT022}) to pre-train our model on $\mathcal{D}$. Given an image-text example $x=$($\texttt{img}, \texttt{txt}$) from $\mathcal{D}$,
we randomly sample a prefix length $T_{p}$. We feed $x_{<T_p}$ that contains the text prefix and image to the model as input and our objective is to generate $x_{\geq T_p}$ containing the rest of the text as output. The training goal is to condition on $x_{<T_{p}}$ and autoregressively generate the remaining text sequence $x_{\geq T_{p}}$:
{\small
\begin{align}
    \mathcal{L}_{\text{PrefixLM}} = & -\mathbb{E}_{x \sim \mathcal{D}}  \big[ {\log p(x_{\geq T_{p}} \mid x_{<T_{p}})}  \big] \\
   = &  -\mathbb{E}_{x \sim \mathcal{D}} \Big[ \sum_{i \geq T_{p}} {\log p(x_i \mid x_{<i})  } \Big]. \nonumber  
\end{align}
}

\textbf{Warm Starting the Model} In order to pre-train all components of our model end-to-end, we need to warm start the retriever at a good state. Otherwise, if starting with random weights, the retriever would often return irrelevant memory items that would never generate useful training signals.

To avoid this cold-start problem, we propose to construct an initial retrieval dataset with pseudo ground-truth knowledge to give the pre-training a reasonable head start. We create a modified version of the Wikipedia-Image-Text (WIT)~\cite{DBLP:conf/sigir/Srinivasan0CBN21} dataset for this purpose. Each image-caption pair in WIT also comes with a corresponding Wikipedia passage (words surrounding the text). We put together the surrounding passage with the query image and use it as the pseudo ground-truth knowledge that corresponds to the input query. As the passage provides rich information about the image and caption, it definitely is useful for initializing the model. To avoid the model from relying on low-level image features for retrieval, we apply random data augmentation to the input query image.  Given this modified dataset that contains pseudo retrieval ground-truth, we train the query and memory key embeddings by optimizing the following contrastive loss: 
\begin{align*}
\mathcal{L}_{contra} = - \text{logSoftmax}(\texttt{Emb}_{\texttt{Query}}(x)^T \texttt{Emb}_{\texttt{Key}}(\hat{z}))
\end{align*}
where $\hat{z}$ represents the pseudo ground-truth knowledge entry corresponding to the input query $x$.

\begin{table*}[t!]
\centering
\small
    \vspace{-.4in}
\begin{tabular}{l|c|c|c} \toprule
\textbf{VQA Model Name}    &  \textbf{Knowledge Sources} & \textbf{Accuracy ($\%$)} & \textbf{Memory (GB)} \\  \midrule
 MUTAN+AN~\cite{DBLP:conf/cvpr/MarinoRFM19}   & Wikipedia + ConceptNet          & 27.8 & -   \\ 
ConceptBERT~\cite{DBLP:conf/emnlp/GarderesZAL20}   &  Wikipedia          & 33.7 & -   \\
 KRISP~\cite{DBLP:conf/cvpr/MarinoCP0R21}   &  Wikipedia + ConceptNet         & 38.4 & -   \\
 Visual Retriever-Reader~\cite{DBLP:conf/emnlp/LuoZBB21}   &  Google Search          & 39.2 & -    \\
 MAVEx~\cite{DBLP:conf/aaai/WuLSM22}   &  Wikipedia+ConceptNet+Google Images          & 39.4 & -   \\
 KAT-Explicit~\cite{DBLP:conf/naacl/GuiWH0BG22}   &  Wikidata     & 44.3 & \color{green}{1.5} \\
 \midrule
PICa-Base~\cite{DBLP:journals/corr/abs-2109-05014}   & Frozen GPT-3     & 43.3 & \color{blue}{350}   \\ 
PICa-Full~\cite{DBLP:journals/corr/abs-2109-05014}     &  Frozen GPT-3          & 48.0 & \color{blue}{350}    \\
KAT~\cite{DBLP:conf/naacl/GuiWH0BG22} (Single)   &  Wikidata + Frozen GPT-3        & 53.1 & \color{green}{1.5} + \color{blue}{352}+ \color{red}{500}   \\
KAT~\cite{DBLP:conf/naacl/GuiWH0BG22}  (Ensemble)   &  Wikidata + Frozen GPT-3         & 54.4 & \color{green}{4.6} + \color{blue}{352} + \color{red}{500}   \\
ReVIVE~\cite{DBLP:journals/corr/abs-2206-01201} (Single)   &  Wikidata + Frozen GPT-3       & 56.6 & \color{green}{1.5} + \color{blue}{354} + \color{red}{500}   \\ 
ReVIVE~\cite{DBLP:journals/corr/abs-2206-01201} (Ensemble)   &  Wikidata+Frozen GPT-3        & 58.0 & \color{green}{4.6} + \color{blue}{354} + \color{red}{500}\\ \midrule
\method-Base   &  WIT + CC12M + Wikidata + VQA-2      & 55.2 & \color{green}{0.8} + \color{blue}{7.5} + \color{red}{744}  \\
\method-Large   &  WIT + CC12M + Wikidata + VQA-2      & 58.0 & \color{green}{2.8} + \color{blue}{10} + \color{red}{993}  \\
\method   &  WIT + CC12M + Wikidata + VQA-2            & \textbf{59.1} & \color{green}{4.2} + \color{blue}{10} + \color{red}{993}  \\
\bottomrule
\end{tabular}
    \vspace{-.1in}
\caption{\textbf{Visual Question Answering} results on OK-VQA, compared with existing methods that use different knowledge sources. For the memory cost, we assume all models use bfloat16. {\color{green}{Green}} means on-device model parameters that are learnable, {\color{blue}{Blue}} means on-device memory of frozen model parameters, and {\color{red}{Red}} means CPU/disk storage cost that are not involved in computation.}
\label{tab:okvqa}
\end{table*}

\subsection{Knowledge Sources} \label{sec:knowledge-sources}
We use the following four sources of knowledge in our experiments: 
    \textbf{Wikipedia-Image-Text (WIT)}~\cite{DBLP:conf/sigir/Srinivasan0CBN21} consists of the images in Wikipedia, as well as their alt-text captions and contextualized text passages.
    \textbf{Conceptual (CC12M)}~\cite{changpinyo2021cc12m} contains web images paired with alt-text captions. It includes many long-tail entities.
    \textbf{VQA-v2}~\cite{DBLP:journals/ijcv/GoyalKASBP19} is a visual question answering dataset. We merge all question-answer pairs per image into a single passage.
     \textbf{WikiData}~\cite{DBLP:journals/cacm/VrandecicK14} is a structural knowledge graph encoding relations between Wikipedia entities. We linearize all relational triplets per entity into a textual passage following the procedure of~\cite{DBLP:journals/corr/abs-2012-14610}.
We have listed the statistical details of these knowledge sources in Table~\ref{tab:kb}.

\subsection{Implementation Details} \label{sec:impl}
Incorporating all the components introduced above, \method can be directly pre-trained over large-scale image caption datasets after proper initialization. As our model architecture is based on T5 and ViT, we use pre-trained ViT checkpoints from \cite{DBLP:conf/cvpr/Zhai0HB22} and pre-trained T5 checkpoints from \cite{DBLP:journals/jmlr/RaffelSRLNMZLL20} to initialize the encoder parameters. 
The query head, key head and attentive fusion layers are initialized from upper T5, while the base text encoder is initialized from lower T5.
The combination of these modules can be found in Table~\ref{tab:variant} for three three model variants, \method-Base, \method-Large and \method, of which the largest \method model has around 2 billion parameters. 

\textbf{Distributed Online Retrieval.}
Finding the top-k most-relevant knowledge entries is a standard Maximum Inner Product Search (MIPS) problem. There are approximate search algorithms~\cite{DBLP:conf/nips/Shrivastava014, DBLP:journals/corr/abs-2206-14286} that scale sub-linearly with the size of the knowledge corpus $|C|$. We use TPU-KNN~\cite{DBLP:journals/corr/abs-2206-14286} to conduct distributed MIPS search, by splitting and storing the memory embeddings across all training devices. The query is synced to each device, which retrieves approximate top-K results from its own memory. Then these results are combined to compute the global top-K retrieved items. 

\textbf{Pre-Training Pipeline.}
We first train the multimodal retriever on our modified version of the Wikipedia Image Text (WIT) dataset via $\mathcal{L}_{contra}$. We use the Adafactor optimizer without momentum ($\beta_1=0$, $\beta_2=0.999$), with weight decay of 0.001\footnote{The remaining experiments use the same optimizer configuration.}, and with a peak learning rate of $6e4$, to train for 10 epochs. We use this checkpoint to warm-start our generative pre-training.
 We set the number of retrieved knowledge entries as $K=10$ during pre-training, and use adafactor with a peak learning rate of $1e{-3}$ and inverse squared root learning rate scheduler with 10,000 linear warm-up steps. We use $\mathcal{L}_{PrefixLM}$ as the main objective, adding $\mathcal{L}_{contra}$, $\mathcal{L}_{decor}$ and $\mathcal{L}_{align}$ weighted by $0.01$. We use a batch size of 4096 across 256 CloudTPUv4 chips and train for about 5 days.

\section{Experimental Results}\label{sec:experiment}

\begin{table}[t!]
\centering
\small
\begin{tabular}{l|c} \toprule
\textbf{VQA Model Name} & \textbf{Accuracy ($\%$)}\\ \midrule
ViLBERT~\cite{DBLP:conf/nips/LuBPL19}  & 30.6 \\
LXMERT~\cite{DBLP:conf/emnlp/TanB19} & 30.7 \\
ClipCap~\cite{DBLP:journals/corr/abs-2111-09734} & 30.9 \\
KRISP~\cite{DBLP:conf/cvpr/MarinoCP0R21} & 33.7 \\ 
GPV-2~\cite{DBLP:conf/eccv/KamathCGKHK22} & 48.6 \\ \midrule
\method-Base & 50.4  \\
\method-Large& 51.5  \\
\method & \textbf{52.2}  \\\bottomrule
\end{tabular}
    \vspace{-.1in}
\caption{\textbf{Visual Question Answering} results on A-OKVQA.}
\label{tab:aokvqa}
\end{table}

We evaluate our proposed method on knowledge-based VQA in Sec.~\ref{sec:vqa} and image captioning in Sec.~\ref{sec:caption}. We then conduct ablation studies in Sec.~\ref{sec:ablation} to analyze the impact of each model component on overall performance.

\subsection{Evaluating on Knowledge-Based VQA}\label{sec:vqa}
One of the most knowledge intensive visual-language tasks is knowledge-based visual question answering (VQA), exemplified by the OK-VQA~\cite{DBLP:conf/cvpr/MarinoRFM19} and A-OKVQA~\cite{DBLP:journals/corr/abs-2206-01718} benchmarks.
To finetune our pre-trained model on these VQA tasks, we use the same generative objective where the model takes in an image question pair as input and generates the text answer as output. There are a few differences between the fine-tuning and the pre-training stages: 1) we set the number of retrieved knowledge entries to $K=50$, so the model is able to retrieve sufficient supporting evidence; 2) we freeze the whole base V-L encoder to stabilize training; and 3) we use a batch size of 128, with the Adafactor optimizier, a peak learning rate of 1e-4. 
We use the soft VQA accuracy metric~\cite{DBLP:journals/corr/AntolALMBZP15} to evaluate the model's generated answer.

Our results on OKVQA and A-OKVQA datasets are shown in Table~\ref{tab:okvqa} and Table~\ref{tab:aokvqa} respectively. For OKVQA, earlier attempts that incorporate a fixed knowledge retriever report results that are below 45\%.  Recently a series of works utilize large language models (\eg GPT-3) as implicit knowledge sources, which achieve much better performance with the trade-off of a huge computational cost. \method achieves higher performance than those methods without relying on such large language models\footnote{As shown in the last column of Table~\ref{tab:okvqa}, \method stores external knowledge as value embeddings on disk, occupying 993GB of space. The key embeddings consume 10GB space and are kept in TPU memory for fast lookup. On the other hand, KAT and REVIVE need to load the entire 350GB GPT-3 model in the GPU/TPU memory. Furthermore, storing WikiData on disk consumes 500GB of disk memory.}. Compared with the previous state-of-the-art, KAT and ReVIVE, which also utilizes T5-Large as a generator, \method achieves accuracy of 59.1\%, which is +6.0\% higher than the single KAT~\cite{DBLP:conf/naacl/GuiWH0BG22} model and  +2.5\% higher than ReVIVE~\cite{DBLP:journals/corr/abs-2206-01201}.

\begin{figure}[t!]
    \centering
    \vspace{-.1in}
    \includegraphics[width=1.0\columnwidth]{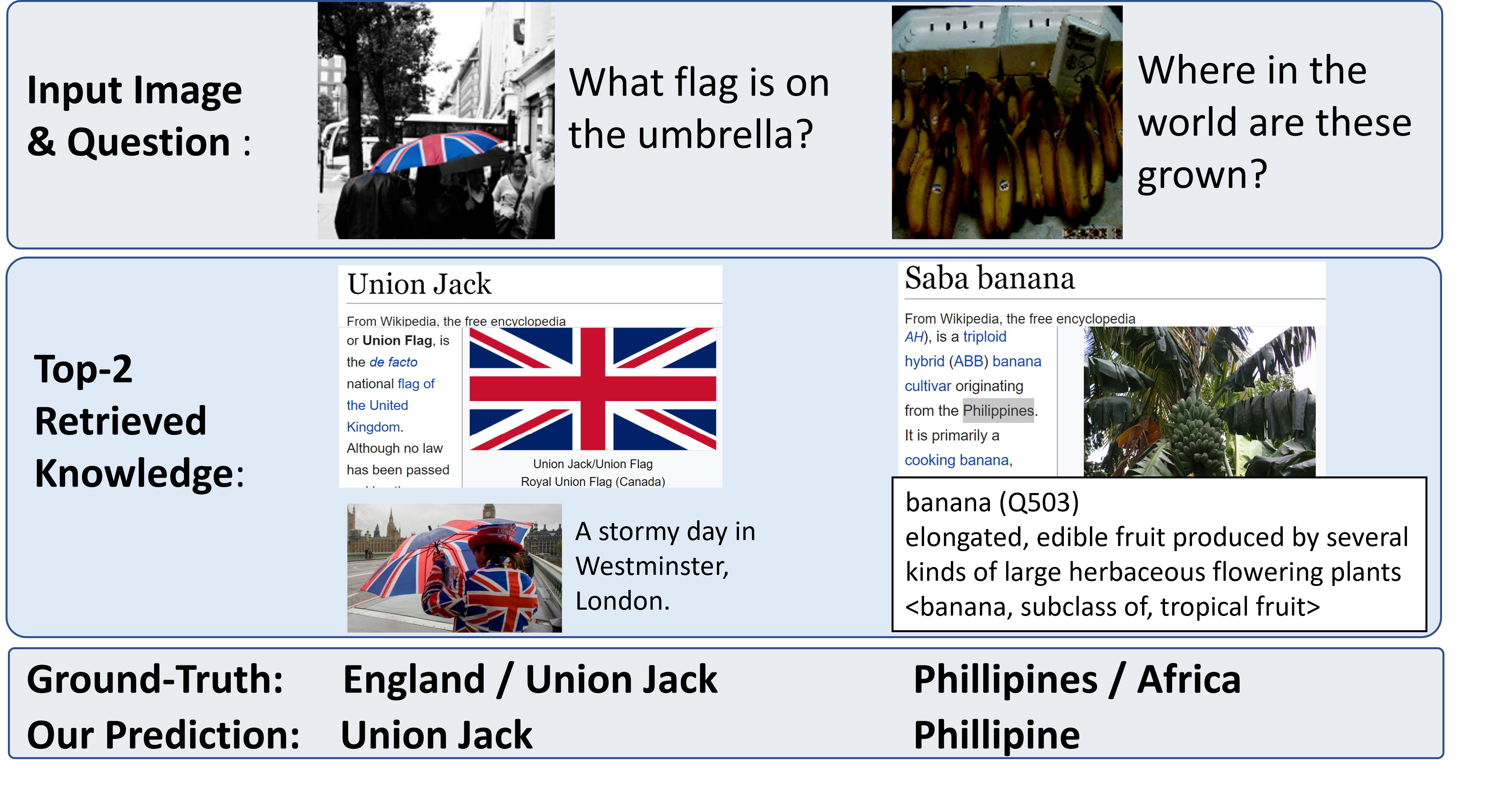}
        \vspace{-.2in}
    \caption{\textbf{VQA Examples}. \method is able to use knowledge from different sources to correctly answer the question. We show more examples in Figure 1-3 of Supplementary Material, indicating that our model can retrieve and use items from diverse knowledge sources to correctly solve different input query.}
    \label{fig:example}
\end{figure}

On A-OKVQA, 
\method achieves 52.2\% accuracy, which is +3.6\% higher than the previous best, GPV-2~\cite{DBLP:conf/eccv/KamathCGKHK22}. We also show two examples of these datasets in Figure~\ref{fig:example}.
All these results show that, with proper end-to-end retrieval training and a diverse set of knowledge sources, \method can learn to retrieve meaningful knowledge entries, and achieve promising results without relying on a large language model.

\subsection{Evaluating on Image Captioning}\label{sec:caption}

We also evaluate \method on image captioning benchmarks: MSCOCO Captions~\cite{DBLP:journals/corr/ChenFLVGDZ15} and NoCaps~\cite{DBLP:conf/iccv/AgrawalAD0CJ0BP19}. We follow the evaluation protocol used in~\cite{DBLP:journals/corr/abs-2205-01917}. We directly fine-tune our generator model on the MSCOCO training split via cross-entropy generative objective. We measure our performance on the MSCOCO test split and NoCaps val set with the CIDEr metric~\cite{DBLP:conf/cvpr/VedantamZP15}. 
The results of these two datasets are shown in Table~\ref{tab:caption}. Note that \method achieves better results than strong recent baselines such as SimVLM~\cite{DBLP:conf/iclr/WangYYDT022} and CoCa~\cite{DBLP:journals/corr/abs-2205-01917} on both benchmarks. Notably, \method$-{\text{Large}}$ with 1.4B parameters outperforms the 2.1B-parameter CoCa model and is significantly better than 80B-parameter Flamingo model~\cite{DBLP:journals/corr/abs-2204-14198}. 

\begin{table}[t!]
\centering
\footnotesize
\vspace{-.1in}
\begin{tabular}{l|cc|c} \toprule
\textbf{Model Name} & \textbf{MSCOCO} & \textbf{NoCaps} & \textbf{\# params.} \\ \midrule
Flamingo~\cite{DBLP:journals/corr/abs-2204-14198}  & 138.1 & -   & 80B\\
VinVL~\cite{DBLP:conf/cvpr/ZhangLHY0WCG21} & 140.9 & 105.1 & 0.4B \\
SimVLM~\cite{DBLP:conf/iclr/WangYYDT022}  & 143.3 & 112.2 & 1.5B\\
CoCa~\cite{DBLP:journals/corr/abs-2205-01917} & 143.6 & 122.4 & 2.1B \\ 
\midrule
\method-Base & 141.1 & 115.8 & 0.4B  \\
\method-Large & 144.5 & 121.3 & 1.4B \\
\method & \textbf{145.4} & \textbf{123.0} & 2.1B \\\bottomrule
\end{tabular}
    \vspace{-.1in}
\caption{\textbf{Image Captioning} results on MSCOCO (Karpathy-test split) and NoCaps (val set). Evaluated using the CIDEr metric.}
\label{tab:caption}
\end{table}

\subsection{Analyzing Effects of Key Model Components}\label{sec:ablation}
In the following we study which design choices contribute most to the model's performance. We focus on three research questions:
    (1) Does utilizing multiple knowledge sources enhance performance?
    (2) Does the proposed attentive fusion surpass existing end-to-end retrieval training methods?
    (3) Can we add knowledge by only updating the memory without modifying model parameters?

\begin{figure}
\begin{floatrow}
\footnotesize
\ffigbox{
    \includegraphics[width=1.0\columnwidth]{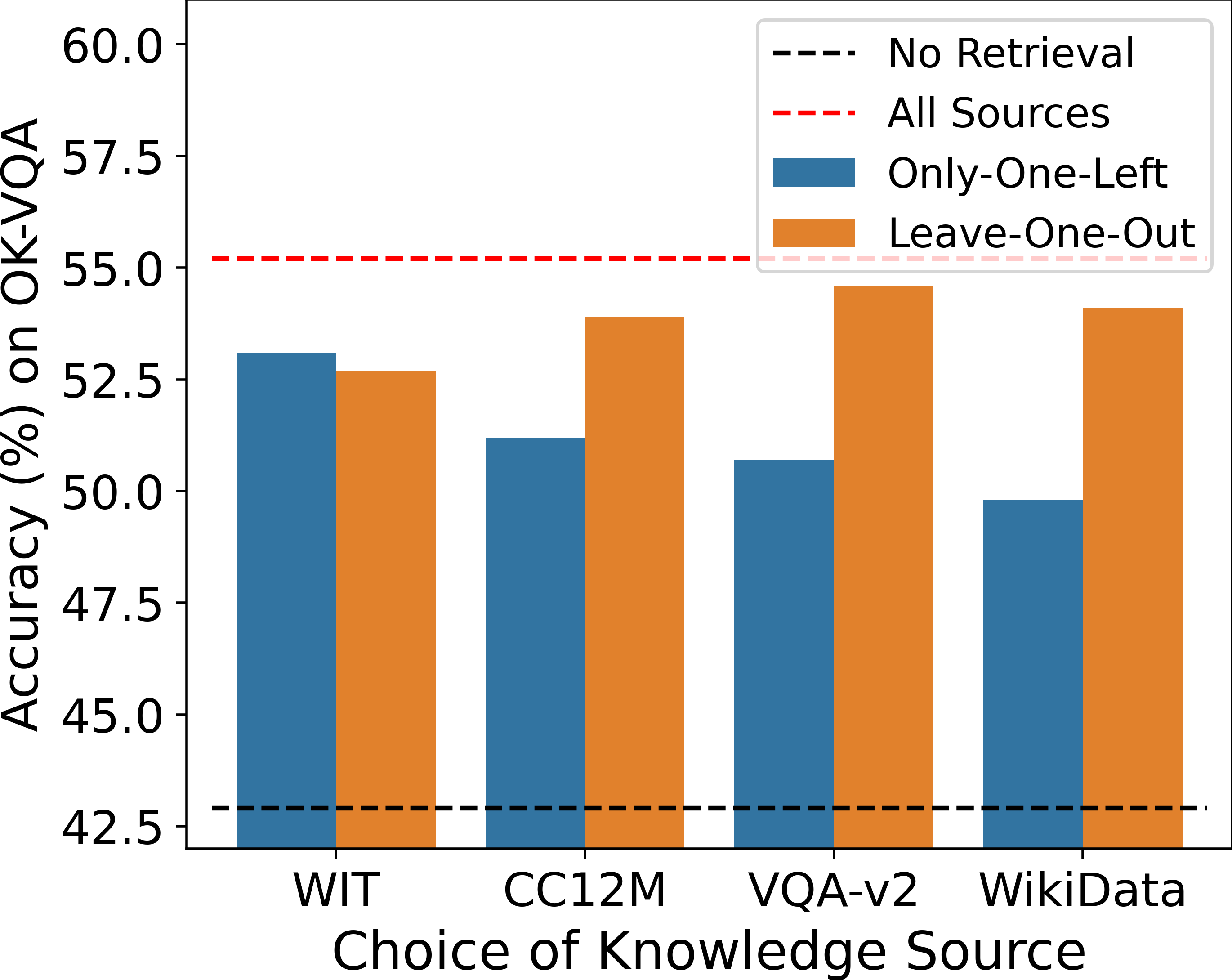}
}{%
\vspace{-.2in}
  \caption{OKVQA Accuracy of \method using 1) \textbf{Only-One-Left}: only use a single knowledge source; 2) \textbf{Leave-One-Out}: use all without this knowledge source.} \label{fig:one}
}
\hspace*{-.15in}
\ffigbox{
\vspace*{-.1in}
  \includegraphics[width=1.1\columnwidth]{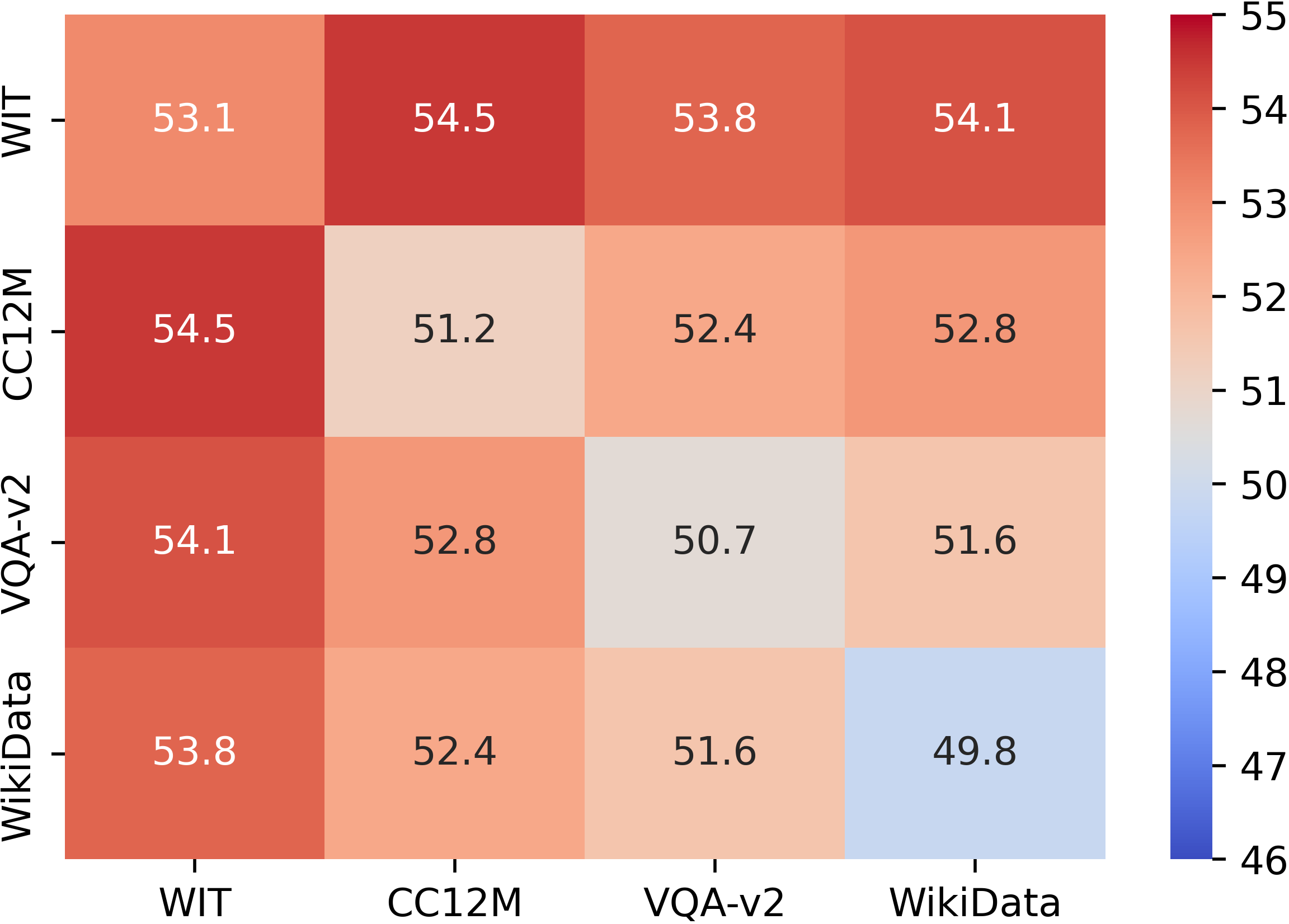}
}{
\vspace{-.2in}
    \caption{OKVQA Accuracy of \method using all \textbf{Pair of Knowledge Sources}. Results show that combining multiple sources could consistently improve performance.} \label{fig:pair}
}

\end{floatrow}
\end{figure}

\textbf{Analyzing multiple knowledge sources.}
A major distinction of \method compared to previous retrieval-augmented approaches is its capacity to utilize a diverse set of knowledge sources during inference. To assess the relative importance of each data source and the efficacy of retrieving from various corpora, we conduct two ablation studies: 1) \textbf{Only-One-Left}: employing a single knowledge source to evaluate the outcomes; and 2) \textbf{Leave-One-Out}: excluding one knowledge source from the complete set $\mathcal{C}$. These ablation studies are executed using the \method$_{\text{Base}}$, evaluated on the OKVQA validation set under the aforementioned conditions.
As shown in Figure~\ref{fig:one}, among the four knowledge sources utilized in this paper, WIT is the most informative, with the highest accuracy when used in isolation (53.1
The remaining three corpora, CC12M, VQA-v2, and WikiData, do not offer the same level of informativeness as WIT when utilized independently. However, excluding any of these corpora from the complete dataset results in performance decreases of 1.3\%, 0.6\%, and 1.1\%, respectively. This observation implies that these knowledge sources effectively complement one another, contributing valuable information to enhance performance. To further substantiate this hypothesis, we perform an additional experiment involving pairs of knowledge sources, as illustrated in Figure~\ref{fig:pair}. Notably, even when paired with an informative knowledge source such as WIT, incorporating an extra corpus consistently leads to performance improvements.

\textbf{Analyzing different retrieval training methods.}
Another core component of \method is the attentive fusion layer, which supports efficient joint training of the retriever and generator. We investigate its performance compared to two existing retrieval training method categories: 1) a frozen retriever based on ALIGN~\cite{DBLP:conf/icml/JiaYXCPPLSLD21} representations ; 2) end-to-end retrieval training methods including \textbf{Attention Distill}~\cite{izacard2020distilling},
 \textbf{EMDR$^2$}~\cite{DBLP:conf/naacl/YangXLLTXLL19},
 and \textbf{Perplexity Distill}~\cite{DBLP:journals/corr/abs-2208-03299}.

We use the pre-trained \method-Base model, fix the generator and randomly initialize the retriever (query head and key head). We utilize our modified version of WIT dataset with pseudo ground-truth retrieved labels as the evaluation corpus. We evaluate retrieval performance by checking whether the correct passage appears in top-10/100 results. For the ALIGN model, we directly evaluate the retrieval results from the pre-trained checkpoint, while for other models, we perform retrieval-augmented training on the WIT dataset. To prevent the model from relying on image similarity for accurate results, we only use text passages as knowledge entries and discard images. Subsequently, we finetune the model on OKVQA and report its accuracy.
The results are presented in Table~\ref{tab:signal}. We observe that directly using pre-trained encoder does not perform well, even with a strong model like ALIGN. 
Moreover, among the various end-to-end retrieval training approaches, our attentive fusion method attains better accuracy in both retrieval and OKVQA tasks. Importantly, our technique exhibits a computational cost (quantified by GFLOPs) comparable to that of attention distillation, yet significantly lower than EMDR$^2$ and Perplexity distillation. This indicates that our proposed method is more efficient and effective for pre-training retrieval-augmented visual-language models.

{ 
\setlength\tabcolsep{2pt} 
\begin{table}[t!]
\centering
\footnotesize
\begin{tabular}{c|cc|c|c} \toprule
\textbf{Retrieval Method} & \textbf{Acc@10} & \textbf{Acc@100} & \textbf{OKVQA Acc.} & \textbf{GFLOPs} \\ \midrule
ALIGN~\cite{DBLP:conf/icml/JiaYXCPPLSLD21} (fixed) & 0.638 & 0.793 & 44.7 & -\\ \midrule
Attention Distill~\cite{izacard2020distilling} & 0.674 & 0.835 & 45.9 & 119\\
EMDR$^2$~\cite{DBLP:conf/naacl/YangXLLTXLL19} & 0.691 & 0.869 & 46.5 & 561\\ 
Perplexity Distill~\cite{DBLP:journals/corr/abs-2208-03299} & 0.704 & 0.886 & 46.7 & 561 \\ \midrule
Ours (Attentive Fusion) & 0.726 & 0.894 & 47.3 & 120 \\
\bottomrule
\end{tabular}
\vspace{-.1in}
\caption{\textbf{Analysis of Retrieval Training Method}: We train \method-Base (frozen generator, only train randomly initialized retriever) to retrieve from the WIT dataset (only text passage without image), and show the retrieval accuracy at the first 10 or 100 results, as well as fine-tuned OKVQA accuracy.}
\label{tab:signal}
\end{table}
}  

\begin{figure}[t!]
    \centering
    \includegraphics[width=0.85\columnwidth]{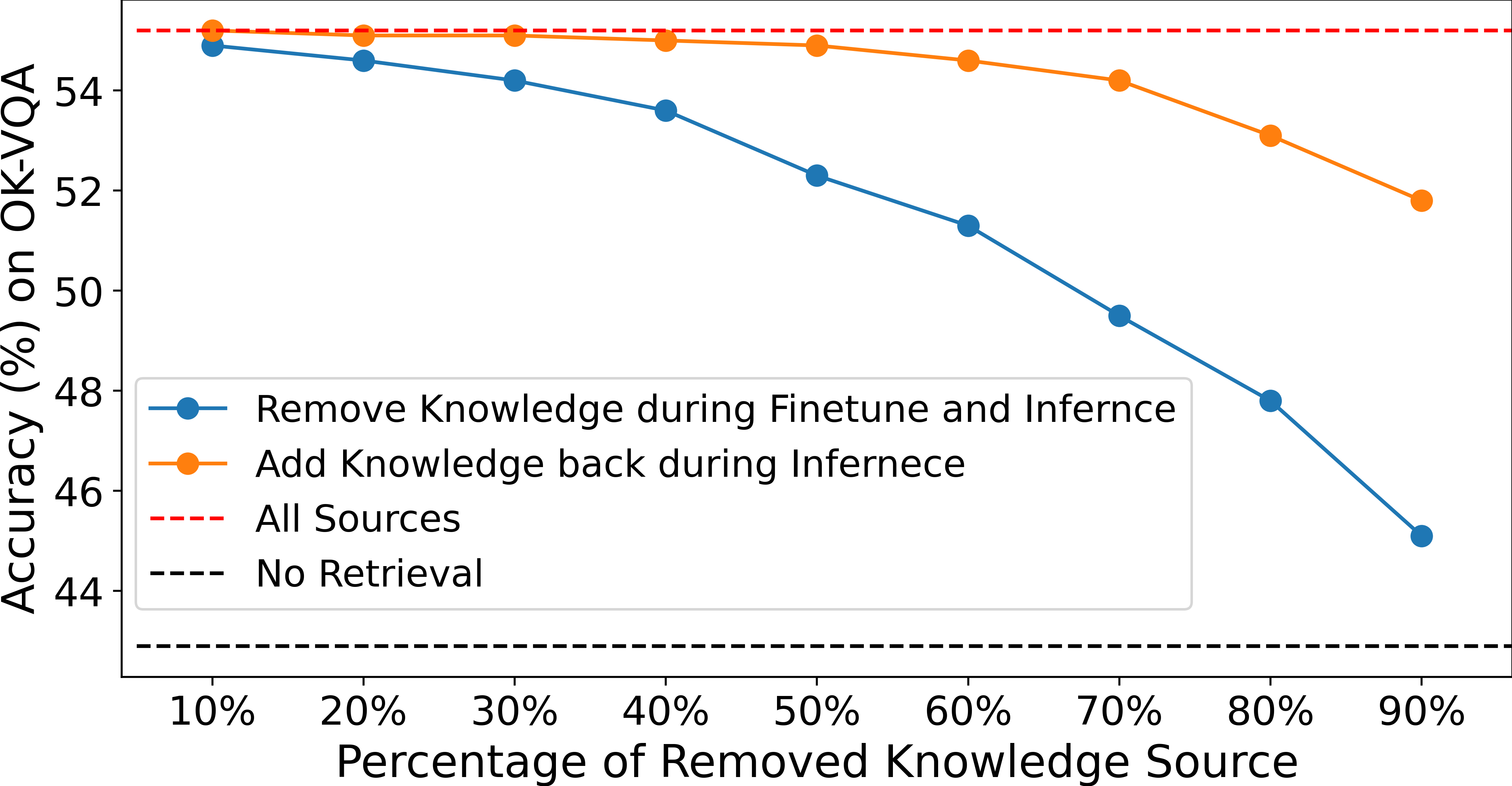}
    \vspace{-.1in}
    \caption{\textbf{Study of Knowledge Update}. The blue curve shows result by removing certain percentage of knowledge during both fine-tuning and inference stage. The orange curve shows results by still first removing the knowledge, and then adding the knowledge back during inference, which simulates the knowledge update.}
    \label{fig:update}
\end{figure}


\textbf{Analyzing Knowledge Modification.}
One advantage of utilizing knowledge memory is that we could easily add or update knowledge entries without re-training model's parameters. To validate this, we conducted ablation studies in which we removed a specific percentage of knowledge entries from the corpora and assessed the performance of the \method-Base model on the OKVQA dataset. Subsequently, we add the removed knowledge back into the corpora, allowing the trained model to make predictions using the complete set of corpora. This approach ensured that the removed knowledge was not seen by the model during fine-tuning, enabling us to test its ability to accurately retrieve and utilize that knowledge for problem-solving.

The results are illustrated in Figure~\ref{fig:update}, with the blue curves representing the inference outcomes without the removed knowledge and the orange curve depicting the results after adding the removed knowledge back. 
A notable performance improvement was observed upon reintroducing the knowledge (orange curve) compared to the outcomes with the removed knowledge (blue curve). Specifically, for the model fine-tuned with only 10\% of the knowledge, the reintroduction of the removed knowledge resulted in an accuracy of 51.8 (+6.7 higher than when removed). This finding demonstrates that the \method model can swiftly adapt to new knowledge by merely updating the memory, obviating the need for re-training model parameters.







\section{Conclusion}\label{sec:conclusion}

This paper presents an end-to-end \underline{Re}trieval-augmented \underline{Vi}sual \underline{L}anguage model (\method), which contains a knowledge retriever that learns to utilize a diverse set of knowledge sources with different modality. The retriever is trained jointly with the generator to return multiple knowledge entries. 
We pre-train \method on a massive image-text corpus with four diverse knowledge corpora, and achieves state-of-the-art results on knowledge-intensive visual question answering and image caption tasks.
In the future we'd explore the ability of this model to be used for attribution, and applying it to broader class of multimodal tasks.


{\small
\bibliographystyle{ieee_fullname}
\bibliography{ref}
}


\end{document}